\documentclass{article}

\usepackage[final]{neurips_2024}

\usepackage{natbib}
\usepackage{caption}
\usepackage{subcaption}
\usepackage{graphicx}
\usepackage{booktabs}
\usepackage{amsfonts, amsmath}
\usepackage[colorlinks=true, citecolor=blue, linkcolor=blue, urlcolor=blue]{hyperref}

\title{Language translation, and change of accent for speech-to-speech task using diffusion model}

\author{%
  Abhishek Mishra\\
  CMINDS, IIT Bombay\\
  abhishek@minds.iitb.ac.in
  \And
  Ritesh Sur Chowdhury\\
  CMINDS, IIT Bombay\\
  24m2154@iitb.ac.in
  \And
  Vartul Bahuguna\\
  CSE, IIT Bombay\\
  24m0828@iitb.ac.in
  \And
  Isha Pandey\\
  CSE, IIT Bombay\\
  iishapandey@iitb.ac.in
  \And
  Ganesh Ramakrishnan\\
  CSE, IIT Bombay\\
  ganesh@cse.iitb.ac.in
  \And
}

\begin{document}

\maketitle

\begin{abstract}
 Speech-to-speech translation (S2ST) aims to convert spoken input in one language to spoken output in another, typically focusing on either language translation or accent adaptation. However, effective cross-cultural communication requires handling both aspects simultaneously — translating content while adapting the speaker’s accent to match the target language context. In this work, we propose a unified approach for simultaneous speech translation and change of accent, a task that remains underexplored in current literature. Our method reformulates the problem as a conditional generation task, where target speech is generated based on  phonemes and guided by target speech features. Leveraging the power of diffusion models, known for high-fidelity generative capabilities, we adapt text-to-image diffusion strategies by conditioning on source speech transcriptions and generating Mel spectrograms representing the target speech with desired linguistic and accentual attributes. This integrated framework enables joint optimization of translation and accent adaptation, offering a more parameter-efficient and effective model compared to traditional pipelines. 
\end{abstract}

\section{Introduction}

Speech-to-Speech Translation (S2ST) has emerged as a critical task in enabling seamless cross-lingual communication. Traditional S2ST systems decompose the task into a sequence of Automatic Speech Recognition (ASR), Machine Translation (MT), and Text-to-Speech (TTS) synthesis modules. While this pipeline architecture is modular and interpretable, it is hindered by compounding errors between components, increased latency, and difficulty in preserving paralinguistic features such as speaker identity and prosody \cite{jia2019translatotron,voicebox2024}.

To address these limitations, recent research has shifted towards end-to-end (E2E) architectures that aim to directly map source speech to target speech. Translatotron \cite{jia2019translatotron} proposed a sequence-to-sequence model for direct speech-to-speech translation without intermediate textual representations. Its successor, Translatotron 2 \cite{jia2022translatotron2}, introduced a modular E2E model that decouples the speech translation and synthesis processes while still maintaining a unified training objective. It significantly improves intelligibility and speaker similarity through a combination of improved target-speaker embeddings and attention-based decoding.

Another line of research focuses on speech-to-unit translation, wherein models map source speech into a sequence of discrete acoustic units—representing phonetic or semantic content—which are then converted into target speech. This paradigm enables disentanglement of linguistic and speaker-specific information, enhancing transferability and reducing data requirements. VioLA \cite{bansal2024viola} exemplifies this by introducing a Voice-to-Language Aligner that translates between speech and multilingual discrete units while preserving speaker and prosodic features. Such models typically rely on self-supervised representation learning techniques like HuBERT \cite{hsu2021hubert} and wav2vec 2.0 \cite{baevski2020wav2vec2}, which have demonstrated the ability to extract high-quality latent features from large-scale unlabeled speech corpora.

The introduction of large-scale multilingual corpora has also spurred progress in speech translation. GigaST \cite{chen2024gigast}, for example, combines over 10 billion text pairs with 8 million speech-text pairs across 38 languages to train robust S2ST systems using joint pretraining strategies. It integrates multilingual speech representation models such as XLS-R \cite{babu2021xlsr} and builds upon architectures like mBART \cite{liu2020mbart} and NLLB \cite{nllb2022} for machine translation. The unified encoder-decoder framework enables transfer learning across multiple languages and modalities, improving generalization to low-resource settings.

At the speech synthesis end, discrete unit-based TTS models have emerged as the backbone of modern S2ST systems. VALL-E X \cite{wang2024vallex} extends the zero-shot capabilities of VALL-E by using quantized speech tokens generated by neural codecs such as EnCodec \cite{defossez2022encodec}. This enables high-quality cross-lingual voice synthesis while preserving speaker identity and prosody from a single demonstration. Meta’s Voicebox \cite{voicebox2024} advances this further through a non-autoregressive speech generation architecture that supports speech editing, infilling, and multilingual synthesis using a large corpus of speech-unit-text triplets. By leveraging Flow Matching \cite{lipman2022flowmatching} for efficient training and inference, Voicebox achieves both flexibility and quality across a diverse set of speech generation tasks.

Despite these breakthroughs, several open challenges remain. Maintaining expressive prosody, enabling zero-shot speaker and language transfer, and scaling models to cover long-tail, under-resourced languages demand further innovation. Furthermore, reducing latency and computational overhead remains essential for real-time applications. The continued integration of self-supervised learning, multilingual pretraining, and discrete unit modeling is likely to define the next generation of universal and efficient S2ST systems.

\begin{figure}
    \centering
    \includegraphics[width=5in]{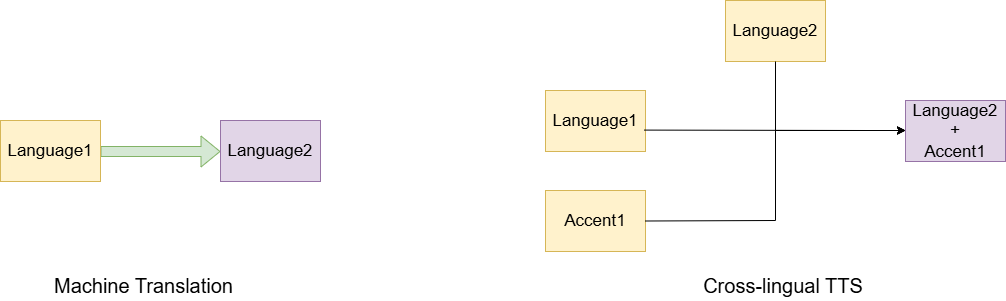}
    \caption{Schematic of existing works}
    \label{existing}
\end{figure}

\begin{figure}
    \centering
    \includegraphics[width=5in]{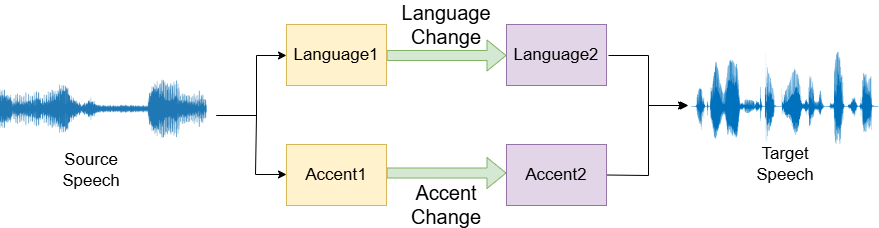}
    \caption{Problem statement of our task}
    \label{problem_statement}
\end{figure}

Previous works in speech mostly focus on introducing accent to form accented speech. Several works have been done on cross-lingual TTS as well. On the other hand, machine translation focuses on changing the language. Figure \ref{existing} shows the existing work done in the previous literature. However, we focus on a more challenging task. Foreign languages and accents are very different and understanding any one of these may be challenging for anyone. To establish effective communication, one must not only translate the language, but also adapt the accent. Thus, our problem is to model an optimal model which can both translate and change the accent from a source speech to a target speech. Figure \ref{problem_statement} shows the problem statement of our work. 

In this work, we have proposed a diffusion based approach to handle this task. First, we have shown that language translation and accent change can be framed as a conditional text-to-image generation task. Then, we have used a diffusion model for solving this task. We have used different sub-tasks for TTS, and S2ST, enabling fine-grained control and better adaptability across languages and accents.. The contributions of this work is summarized as follows:

\begin{itemize}
    \item We introduced a novel way of modeling language translation and accent change using conditional diffusion models.

    \item We develop a pipeline utilizing diffusion models for different TTS and S2ST tasks.

    \item We empirically validate the effectiveness of our method using benchmark datasets, showing improved audio quality.

    \item We provide an extensible framework that can be adapted for multilingual and multi-accent speech generation tasks.
\end{itemize}

This paper is structured as follows. Section 2 describes the related works. Section 3 presents the methodology. Section 4 presents the summary of the tasks we used. Section 5 and 6 present the implementation details and the experimental results. Finally the conclusion and future works have been discussed in section 7 and 8 respectively.

\section{Related Work}
\begin{itemize}
    \item \textbf{Cross-Lingual TTS} -  In cross-lingual speech synthesis, the goal is to synthesize the speech of
 another language for a monolingual speaker, which is more challenging than conventional monolingual TTS[\cite{chorowski2019unsupervised},\cite{zhang2023multilingual},\cite{yang2022cross},\cite{cai2023cross}].By using shared phonemic input representation across languages and incorporating an adversarial
 objective to disentangle the speaker’s identity and speech content,\cite{ren2019fastspeech} is able to
 achieve cross-lingual voice cloning within limited speakers.Zero shot TTS has also been achieved by prompting with speaker voice[\cite{zhang2023speak}].In this system the achieved they achieved it using Auto Regressive and Non Auto Regressive Codec Language Model on Chinese and English Languages.Diffusion based tts has also been came up with the idea where you give any text Diff-TTS[\cite{jeong2021diff}]exploits a denoising diffusion framework to transform
the noise signal into a mel-spectrogram via diffusion time steps which then with the help of encoder and decoder converts it into speech.

    \item \textbf{Speech-to-speech Translation (S2ST)} -  S2ST aims to translate the speech of one language to the
 speech of another language . The initial research and application mainly focus on cascaded S2ST
 systems [\cite{nakamura2006atr},\cite{lavie1997janus}].[\cite{zhang2023speak}] Also came up with s2s translation where they have used an additional speech recognition and translation model for translation and extraction of phhonemes.
 here has also been growing interest in multilingual S2ST systems and methods for low-resource languages. Facebook’s M2M-100 [\cite{fan2021beyond}] and Meta’s SeamlessM4T [\cite{barrault2023seamlessm4t}] tackle multilingual speech translation using shared encoders and decoders, leveraging massive multilingual corpora. Techniques like transfer learning, self-supervised pretraining (e.g., wav2vec 2.0), and data augmentation have also been applied to improve performance in data-scarce settings .
    
    \item \textbf{Diffusion} - Several recent works propose different aspects of issues on diffusion models. One such issue is high computational cost incurred by repetitive calling of the neural network model. Latent diffusion models \cite{rombach2022high} solve this issue by compressing irrelevant perceptual details. DDIM (Denoising Diffusion Implicit Models) \cite{song2020denoising} generalizes the DDPM in non-markovian structure leading to deterministic sampling and faster generation. Several works \cite{nichol2021improved} are based on improving variance scheduling like cosine, sigmoid etc. and including different type of loss functions and regularizations. Guided diffusion \cite{song2020score} includes classifier/classifier-free guidance, which helps to includes classes in the diffusion model. ControlNet \cite{zhang2023adding} introduces detailed conditioning (e.g., text, edges, segmentation, texture) for controllable generation.

\end{itemize}

\section{Methodology}

\subsection{Phoneme-Conditional Speech Generation for Speech-to-Speech Tasks}
\label{theory}

We propose that our task is a conditional generation task, where we generate the target speech conditioned by the source phonemes.

Suppose, $p_s, a_s, p_t$ and $a_t$ are the source phoneme, acoustic and target phoneme, acoustic respectively. Let's consider $f_s(p_s,a_s)$ and $f_t(p_t,a_t)$ are the source and target speech respectively. 

Then, the conditional distribution of generating target speech w.r.t. source speech is given by $P(f_t(p_t, a_t) \mid f_s(p_s, a_s))$. Using law of total probability, we get,

\begin{equation}
P(f_t(p_t, a_t) \mid f_s(p_s, a_s)) = \sum\limits_{(p_s, a_s)} P(f_t(p_t, a_t) \mid p_s, a_s) P(p_s, a_s \mid f_s(p_s, a_s))
\end{equation}

Now, our assumption is that, target phonemes and acoustic is independent of source acoustic given source phoneme, thus, $(p_t, a_t) \perp a_s \mid p_s$. Thus,

\begin{equation}
P(f_t(p_t, a_t) \mid f_s(p_s, a_s)) = \sum\limits_{(p_s, a_s)} P(f_t(p_t, a_t) \mid p_s) P(p_s, a_s \mid f_s(p_s, a_s))
\end{equation}

As, $P(f_t(p_t, a_t) \mid p_s)$ is not dependent of $a_s$, we get,
\begin{equation}
P(f_t(p_t, a_t) \mid f_s(p_s, a_s)) = \sum\limits_{p_s} \left[ P(f_t(p_t, a_t) \mid p_s) \sum\limits_{a_s} P(p_s, a_s \mid f_s(p_s, a_s)) \right] 
\end{equation}

Marginalizing over $a_s$, we get,
\begin{equation}
P(f_t(p_t, a_t) \mid f_s(p_s, a_s)) = \sum\limits_{p_s} P(f_t(p_t, a_t) \mid p_s) P(p_s \mid f_s(p_s, a_s))
\end{equation}
Now, we can construct $f_s'(.)$, such that,
\begin{equation}
P(f_t(p_t, a_t) \mid f_s(p_s, a_s)) = \sum\limits_{p_s} P(f_t(p_t, a_t) \mid p_s) P(p_s \mid f_s'(p_s))
\end{equation}
Again, using law of total probability, we get,
\begin{equation}
P(f_t(p_t, a_t) \mid f_s(p_s, a_s)) = P(f_t(p_t, a_t) \mid f_s'(p_s))
\end{equation}

Thus, based on our assumption, we can consider the language translation and accent change as a conditional generation task, where the target speech is generated conditioned on source phonemes.

\subsection{Basics of Diffusion Probabilistic Models}

Diffusion based models are the state-of-the-art generative models. It offers high quality generation, as well as avoiding mode collapse. In training stage, diffusion process adds noise in forward process and denoise the noisy data using neural network in small steps.

\subsubsection{Forward Diffusion}

The forward diffusion process for stochastic process $X_t$ is defined by the stochastic differential equation (SDE):
\begin{equation}
    dX_t = -\frac{1}{2} \Sigma^{-1}(X_t - \mu)\beta_t \, dt + \sqrt{\beta_t} \, dW_t, \quad t \in [0, T]
\end{equation}

Where $T$ is the number of timestep, $\beta_t$ is the noise schedule which indicates the amount of noise added, $W_t$ is the gaussian noise, $\mu$ is a vector, and $\Sigma$ is a positive diagonal matrix.

The solution to this SDE is:
\begin{equation}
    X_t =  e^{-\frac{1}{2} \Sigma^{-1} \int_0^t \beta_s \, ds} X_0 + \left(I - e^{-\frac{1}{2} \Sigma^{-1} \int_0^t \beta_s \, ds} \right)\mu + \int_0^t \sqrt{\beta_s} \, e^{-\frac{1}{2} \Sigma^{-1} \int_s^t \beta_u \, du} \, dW_s
\end{equation}

Using Ito's integral, it can be shown that, for $T \rightarrow \infty$ and $\lim_{T \rightarrow \infty} e^{-\int_0^T \beta_s ds} = 0$, the conditional distribution of $X_T$ given $X_0$ is given as:

\begin{equation}
    P(X_T \mid X_0) = \mathcal{N}(\mu, \Sigma)
\end{equation}

Which signifies that, the conditional distribution of $X_T$ given $X_0$ approaches Normal distribution with mean $\mu$ and covariance $\Sigma$.

\subsubsection{Reverse Diffusion}

The score-based reverse-time SDE is given by:

\begin{equation}
    dX_t = - \frac{1}{2} \Sigma^{-1} (X_t - \mu)\beta_t dt - \nabla \log p_t(X_t)  \beta_t dt + \sqrt{\beta_t}  d\widetilde{W}_t, \quad t \in [0, T]
\end{equation}

Where, $\nabla \log p_t(X_t)$ and $d\widetilde{W}_t$ denote the score function and reverse-time Brownian motion respectively.

To solve this equation, we train a neural network that approximates the score function using equation:

\begin{equation}
    \min_{\theta} \mathbb{E}_{t \sim \mathcal{U}(0, T)} \mathbb{E}_{\mathbf{x}_t \sim p_t(\mathbf{x}_t)} \left\| \mathbf{s}_\theta(\mathbf{x}_t, t) - \nabla_{\mathbf{x}_t} \log p_t(\mathbf{x}_t) \right\|_2^2
\end{equation}

\subsection{Model details}
In this work, we have used the GradTTS \cite{popov2021grad} model as our primary model. In addition, we have used multiple adapters to perform different tasks. Grad-TTS is a diffusion-based, non-autoregressive text-to-speech model designed to synthesize high-quality and expressive speech from text inputs. The architecture, as shown in figure \ref{gradtts}, consists of three primary stages: text encoding and alignment, score-based decoding, and reverse diffusion synthesis.

\begin{itemize}
    \item \textbf{Architecture:}
    
    The process begins with a phoneme sequence derived from the input text. This sequence is passed through the encoder, which transforms each phoneme into a corresponding latent representation, denoted as $\tilde{\mu}$. Simultaneously, a duration predictor estimates how many acoustic frames each phoneme should span. These predicted durations are used to expand or repeat the encoded phoneme representations accordingly. The result is an aligned latent representation $\mu$, which now matches the temporal resolution of the target mel-spectrogram frames.

Next, this aligned representation $\mu$ is fed into a score-based generative model that performs reverse diffusion through an ordinary differential equation (ODE) solver. The generative process starts with a noisy latent variable $X_t$, sampled from a Gaussian distribution centered at $\mu$, specifically $X_T \sim \mathcal{N}(\mu, I)$. A U-Net, conditioned on the aligned latent representation $\mu$ and the current timestep $t$, estimates the score function $s_\theta(X_t, \mu, t)$, which approximates the gradient of the data distribution's log-probability. This score guides the reverse diffusion process by iteratively removing noise from the initial sample.

Finally, the generated mel-spectrogram $X_0$ is passed to a neural vocoder, such as HiFi-GAN, which synthesizes the final waveform. By combining duration-aware alignment with diffusion-based generation, Grad-TTS achieves fast, high-fidelity, and expressive speech synthesis while allowing for diversity and control through the stochastic nature of its sampling process.

\begin{figure}
    \centering
    \includegraphics[width=5in]{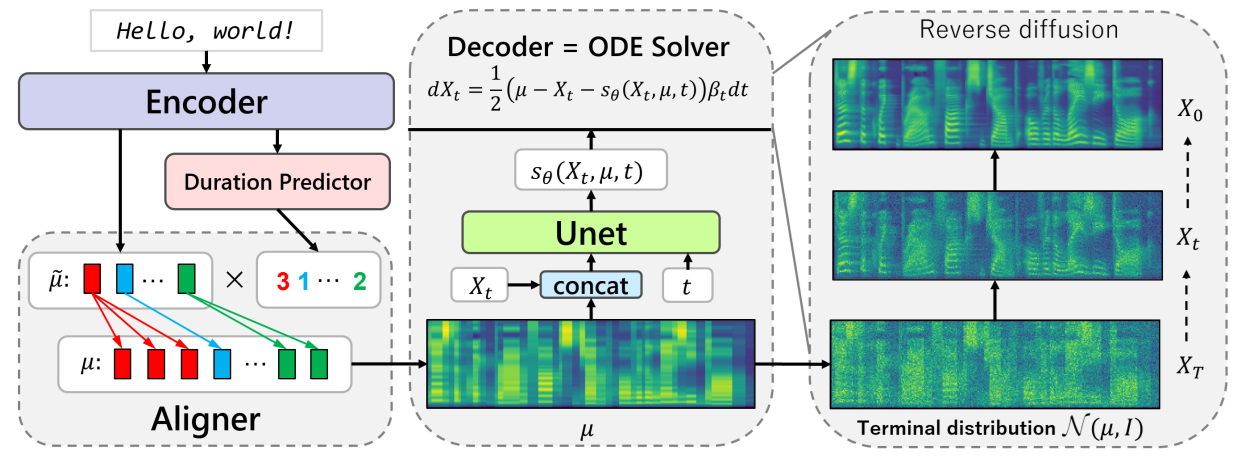}
    \caption{GradTTS Architecture}
    \label{gradtts}
\end{figure}

    \item \textbf{Training procedure:}
    
    To force the Aligner output with distribution $\mu$, we train the encoded output as:
    \begin{equation}
        \mathcal{L}_{enc} = -\sum^F_{j=1} log (\phi(y_j ; \tilde{\mu}_{A(j)}, I))
    \end{equation}
    Where $F$ is the total number of spectrogram frames, $\phi(. ; \tilde{\mu}_i, I)$ is the probability density function of $\mathcal{N}( \tilde{\mu}_i, I)$.

    To obtain the alignment $A^*$ during inference, Grad-TTS utilizes a duration predictor network, which is trained using the Mean Square Error (MSE) loss in the logarithmic domain \cite{kim2020glow}.

The target duration $d_i$ for each phoneme $i$ is computed as:
\begin{equation}
    d_i = \log \sum_{j=1}^{F} \mathbb{I}_{\{A^*(j) = i\}}, \quad i = 1, \ldots, L,
\end{equation}
Where $L$ is the number of input phonemes, and $\mathbb{I}$ is the indicator function that checks if frame $j$ is aligned to phoneme $i$.

The loss for the duration predictor is then defined as:
\begin{equation}
    \mathcal{L}_{dp} = \mathrm{MSE}(DP(sg(\tilde{\mu})), d),
\end{equation}
Where $DP$ denotes the duration predictor, $\tilde{\mu} = \tilde{\mu}_{1:L}$ is the phoneme-level encoded representation, and $sg[\cdot]$ is the stop-gradient operator. This operator ensures that the encoder parameters remain unaffected by the duration prediction loss $\mathcal{L}_{dp}$ by blocking gradient flow during backpropagation.

The loss function of the reverse diffusion process is given by:

\begin{equation}
\mathcal{L}_{\text{diff}} = \mathbb{E}_{X_0, t} \left[ 
    \lambda_t \, \mathbb{E}_{\xi_t}  
        \left\| s_\theta(X_t, \mu, t) + \frac{\xi_t}{\sqrt{\lambda_t}} \right\|_2^2 
\right]
\end{equation}
Where, $\lambda_t = 1 - e^{- \int_0^t \beta_s \, ds}$ and $\xi_t  \sim \mathcal{N}(0,I)$.

Thus, the training procedure is as follows:

\begin{enumerate}
    \item \textbf{Alignment Estimation:} With the parameters of the encoder, duration predictor, and decoder fixed, the Monotonic Alignment Search (MAS) algorithm is employed to compute the optimal alignment \( A^* \) that minimizes the encoder loss \( L_{\text{enc}} \).
    
    \item \textbf{Parameter Optimization:} Given the fixed alignment \( A^* \), the parameters of the encoder, duration predictor, and decoder are updated by minimizing the composite objective function $L_{\text{total}} = L_{\text{enc}} + L_{\text{dp}} + L_{\text{diff}}$.
    
    \item \textbf{Iterative Refinement:} Steps 1 and 2 are alternated iteratively until convergence is achieved.
\end{enumerate}

\end{itemize}

\begin{figure}
     \centering
     \begin{subfigure}[b]{0.4\textwidth}
         \centering
         \includegraphics[width=\textwidth]{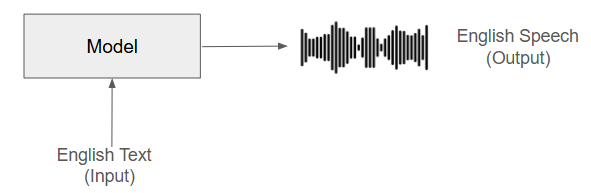}
         \caption{Baseline TTS}
     \end{subfigure}
     \begin{subfigure}[b]{0.4\textwidth}
         \centering
         \includegraphics[width=\textwidth]{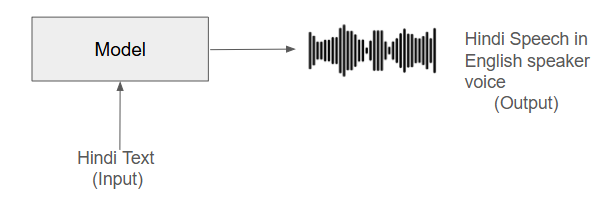}
         \caption{Cross-Lingual TTS}
     \end{subfigure}
     \begin{subfigure}[b]{0.4\textwidth}
         \centering
         \includegraphics[width=\textwidth]{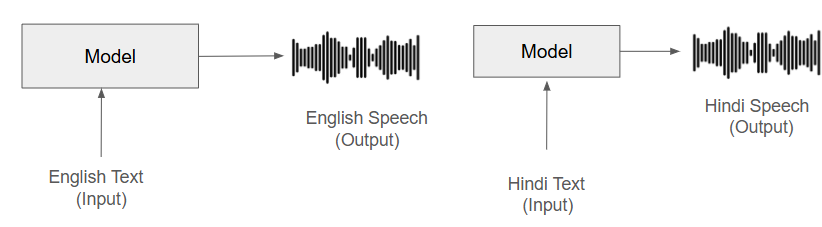}
         \caption{Multi-lingual TTS with random speaker assignment}
     \end{subfigure}
     \begin{subfigure}[b]{0.4\textwidth}
         \centering
         \includegraphics[width=\textwidth]{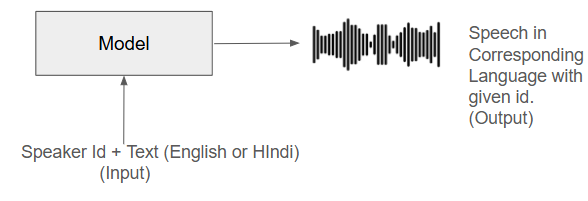}
         \caption{Multilingual TTS with Speaker ID}
     \end{subfigure}
     \begin{subfigure}[b]{0.4\textwidth}
         \centering
         \includegraphics[width=\textwidth]{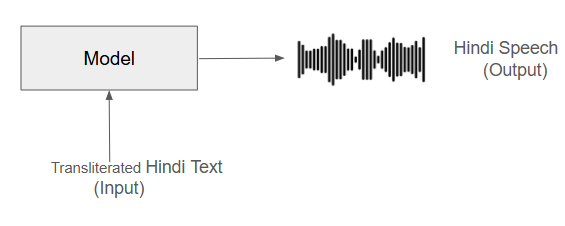}
         \caption{Multilingual TTS with Transliterated Input}
     \end{subfigure}
     \begin{subfigure}[b]{0.4\textwidth}
         \centering
         \includegraphics[width=\textwidth]{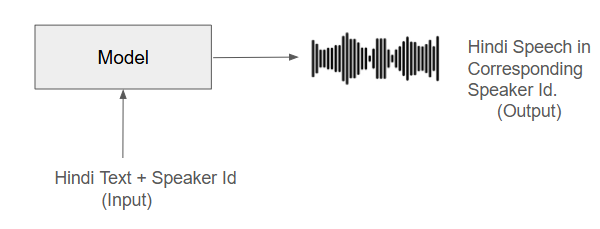}
         \caption{Multilingual TTS with Transliterated Input and Speaker ID Control}
     \end{subfigure}
        \caption{Overview of TTS tasks}
        \label{tasks}
\end{figure}

\section{Summary of tasks}
We have performed several tasks in the model. Figure \ref{tasks} presents the experiments done here.



\subsection{Text-to-Speech}

\begin{itemize}
    \item \textbf{Baseline TTS:} In the baseline TTS task, we have provided english text and generated speech signal with english accent.
    \item \textbf{Cross-Language TTS:} For this task, we have first translated the english text to hindi text and then added english accent to the hindi text. We have trained this task with english text to english speech and inferred with hindi text.
    \item \textbf{Multi-lingual TTS with random speaker assignment:} For this task, we have used english and hindi texts. We have added english/hindi accent to form the speech signal.
    \item \textbf{Multilingual TTS with Speaker ID:} This task is same as Multi-lingual TTS with random speaker assignment task in addition to speaker ID.
    \item \textbf{Multilingual TTS with Transliterated Input:} For this task, we have transliterated the hindi language to english and then introduced english/hindi accent to it.
    \item \textbf{Multilingual TTS with Transliterated Input and Speaker ID Control:} This task is addition of previous two tasks, that is, it takes both transliterated input and speaker ID control.
\end{itemize}

\subsection{Language translation and Accent transfer}

Although we can use the previous models to translate the language and adapt to the target accent, integrating both the tasks into a single model will provide a more parameter-efficient model. Moreover, we can jointly optimize both language translation and accent modification tasks, which helps minimize the disparity between them.

Moreover, we can assume that the accent of the target language is independent of the source language, which makes the accent change a generation task. We have given a mathematical derivation in section \ref{theory}. Thus, target speech generation is a generation task conditioned with target phonemes. We can also use the generative model’s learning in such a way to generate target language by conditioning from source language. Thus, we can utilize the conditional generation models for simultaneous language translation and accent transfer.

Diffusion models offer very high quality generation. Generation with diffusion model will provide better generation than existing works. Text-to-image diffusion models generate images with input prompts. For our task, we can leverage the text-to-image diffusion model with Mel spectrogram of the target speech as targeted output and text prompts of the source speech as conditioning input. Thus, we can use the existing text-to-image diffusion models for solving the task.

\begin{figure}
    \centering
    \includegraphics[width=0.7\linewidth]{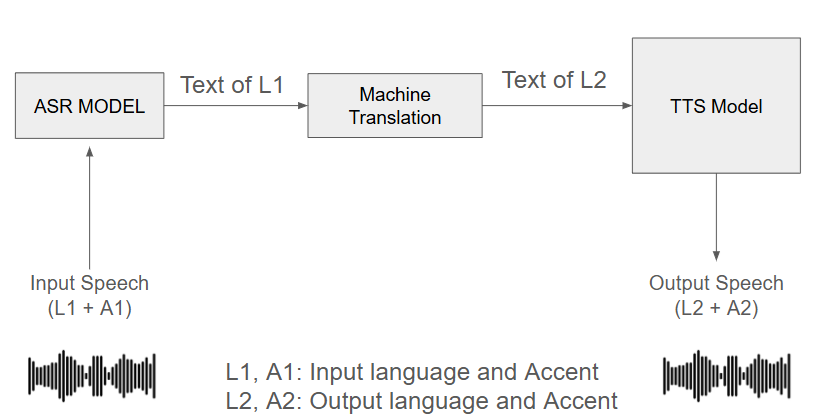}
    \caption{S2ST Pipeline}
    \label{s2st}
\end{figure}

Here, we have used the pipeline shown in figure \ref{s2st}, for this task. We first do the ASR, which converts speech to text; then, machine translation is performed which changes the language, and finally add accent the target accent to the target text, to generate the target speech.

\section{Implementation details}

\subsection{Dataset description}

\begin{itemize}
    \item \textbf{LJSpeech:} The LJSpeech dataset \cite{ljspeech17} is a publicly available dataset for training and evaluating text-to-speech (TTS) systems. It comprises 13,100 English audio clips recorded by a single female speaker, with a total duration of approximately 24 hours. Each clip is paired with its corresponding normalized text transcription. The recordings are sampled at 22.05kHz. All transcriptions are provided in a plain-text format, and the dataset includes a CSV file that maps each audio file to its transcription. The consistent speaking style and high-quality recordings make it well-suited for training TTS models.

    \item \textbf{IndicTTS:} The IndicTTS dataset \cite{indictts} is a multilingual speech dataset developed for building text-to-speech (TTS) systems for Indian languages. It covers 13 major Indian languages including Hindi, Marathi, Bengali, Telugu, Tamil, Gujarati, Kannada, Malayalam, Odia, Punjabi, Urdu, Assamese, and Rajasthani. For each language, the dataset provides high-quality recordings by native speakers. The audio is sampled at 48kHz or 16kHz, depending on the language and speaker. The corpus contains phonetically rich and balanced sentences designed to cover the diverse phoneme distributions of Indian languages. For this work, we have used Hindi language only.

\end{itemize}

\subsection{Evaluation metrics}

In order to assess the performance of the model, we have used the speaker similarity using automatic speaker verification (ASV) model, and word error rate (WER) followed by ASR model, which gives a formalized way of evaluating the performance. Higher ASV score shows the higher similarity between the speech signals, whereas lower ASR-WER suggests the lower error between the ground truth text and the texts generated after performing ASR. Moreover, we have shown the Mean Opinion Score (MOS), which is a human feedback.

\subsection{Experimental details}

We have experimented on NVIDIA GPU H100, which provides 80GB memory. LJSpeech dataset has been divided into train-val-test split as 16hrs-10min-38min and IndicTTS dataset has been split as 17.78hrs-8.25min-29.73min.

For baseline and cross-lingual TTS tasks, we have used the LJSpeech dataset, whereas for multi-lingual tasks, we have used both LJSpeech and IndicTTS datasets.


\section{Experimental Results}

\subsection{TTS}

\begin{table}[h]
\centering
\begin{tabular}{|p{5cm}|c|c|}
\hline
\textbf{Experiment Description} & \textbf{ASV Score} & \textbf{ASR-WER (\%)} \\
\hline
Baseline TTS & 0.8277 & 8.51 \\
\hline
Cross-Language TTS & - & 92.49 \\
\hline
Multilingual TTS with Random Speaker Assignment & English: 0.8101, Hindi: 0.3105 & English: 13.75, Hindi: 6.39 \\
\hline
Multilingual TTS with Speaker ID & English: 0.7992, Hindi: 0.6504 & English: 12.80, Hindi: 23.60 \\
\hline
Multilingual TTS with Transliterated Input & English: 0.8208, Hindi: 0.3678 &  English: 14.25, Hindi: 22.89\\
\hline
Multilingual TTS with Transliterated Input and Speaker ID & English: 0.8160, Hindi: 0.6120 & English: 14.86, Hindi: 26.46 \\
\hline

\end{tabular}
\caption{Performance Metrics of Different TTS Configurations}
\label{tts_results}
\end{table}

Table \ref{tts_results} presents the results over different TTS tasks. The ASV score and ASR-WER are results are very good (ASV: 0.8277, ASR-WER: 8.51\%) for baseline TTS. The cross-lingual TTS shows a poor ASR-WER score (92.49\%), as the hindi tokenizers are unseen to the model. The ASV score is irrelevant here, as the input speech is english and output speech has Hindi text-English accent. 

For multilingual TTS with random speaker assignment, the results are relatively better (For English: ASV score: 0.8101, ASR-WER: 13.75\% and for Hindi: ASV score: 0.3105, ASR-WER: 6.39\%). However, the results for this tasks are not evaluated properly using these metrics, as here speakers are randomly assigned with the speech, for example, Hindi speech can be assigned to an English speaker. We can see an improvement in ASV score with speaker ID in Hindi language, as in IndicTTS dataset there are two Hindi language speakers (male and female). Transliteration has very minor effects on multilingual task.


\subsection{Language Translation and Accent Transfer:} 
    We have generated the translated and target accented speech using pipeline shown in figure \ref{s2st}. However, due to lack of corresponding ground truth, the results can't be presented here. From qualitative perspective, the results has less quality due to cascading errors.

\section{Conclusion}
In this work, we introduced a novel approach to simultaneous language translation and accent adaptation for speech-to-speech tasks using diffusion models. By formulating the problem as a conditional generative task and leveraging the strengths of Grad-TTS and text-to-image style diffusion pipelines, we demonstrated that high-quality and controllable speech synthesis is achievable across multiple languages and accents. Experimental results across various TTS and S2ST settings validate the effectiveness of our approach, highlighting its potential for scalable and expressive cross-lingual communication. This work sets the stage for further exploration into unified, diffusion-based speech generation frameworks for real-world multilingual applications.

\section{Limitations and future works}
While our proposed diffusion-based framework shows promising results in combining language translation with accent adaptation, several limitations remain. First, cascading the models leads to cascading errors. To minimize this error, we will explore methods to combine the ASR, machine translation and TTS into a single model. Second, the computational cost of training and inference with diffusion models is significantly higher compared to traditional autoregressive models, which may hinder real-time deployment. Additionally, the quality of accent transfer can degrade for low-resource languages or unseen accents due to limited data diversity. Our reliance on accurate phoneme extraction and transcription also introduces potential bottlenecks in multilingual scenarios. In future work, we plan to explore lightweight diffusion variants or hybrid models to reduce latency while preserving quality. We also aim to expand the framework to support more language-accent pairs, particularly low-resource combinations, by incorporating self-supervised pretraining and improved alignment mechanisms. Finally, integrating explicit prosody control and speaker disentanglement could enhance expressiveness and personalization in generated speech, pushing us closer to truly universal and adaptive S2ST systems.

\bibliographystyle{plainnat}
\bibliography{main}

\begin{thebibliography}{33}
\providecommand{\natexlab}[1]{#1}
\providecommand{\url}[1]{\texttt{#1}}
\expandafter\ifx\csname urlstyle\endcsname\relax
  \providecommand{\doi}[1]{doi: #1}\else
  \providecommand{\doi}{doi: \begingroup \urlstyle{rm}\Url}\fi

\bibitem[AI(2024)]{voicebox2024}
Meta AI.
\newblock Voicebox: Text-guided multilingual universal speech generation at scale.
\newblock \emph{arXiv preprint arXiv:2404.00569}, 2024.

\bibitem[Babu et~al.(2021)]{babu2021xlsr}
Naveen~Arivazhagan Babu et~al.
\newblock Xls-r: Self-supervised cross-lingual speech representation learning at scale.
\newblock \emph{arXiv preprint arXiv:2111.09296}, 2021.

\bibitem[Baevski et~al.(2020)]{baevski2020wav2vec2}
Alexei Baevski et~al.
\newblock wav2vec 2.0: A framework for self-supervised learning of speech representations.
\newblock \emph{NeurIPS}, 2020.

\bibitem[Bansal et~al.(2024)]{bansal2024viola}
M.~Bansal et~al.
\newblock Viola: Voice-to-language aligner for direct speech-to-speech translation.
\newblock \emph{arXiv preprint arXiv:2503.04799}, 2024.

\bibitem[Barrault et~al.(2023)Barrault, Chung, Meglioli, Dale, Dong, Duquenne, Elsahar, Gong, Heffernan, Hoffman, et~al.]{barrault2023seamlessm4t}
Lo{\"\i}c Barrault, Yu-An Chung, Mariano~Cora Meglioli, David Dale, Ning Dong, Paul-Ambroise Duquenne, Hady Elsahar, Hongyu Gong, Kevin Heffernan, John Hoffman, et~al.
\newblock Seamlessm4t: Massively multilingual \& multimodal machine translation.
\newblock \emph{arXiv preprint arXiv:2308.11596}, 2023.

\bibitem[Cai et~al.(2023)Cai, Yang, and Li]{cai2023cross}
Zexin Cai, Yaogen Yang, and Ming Li.
\newblock Cross-lingual multi-speaker speech synthesis with limited bilingual training data.
\newblock \emph{Computer Speech \& Language}, 77:\penalty0 101427, 2023.

\bibitem[Chen et~al.(2024)]{chen2024gigast}
G.~Chen et~al.
\newblock Gigast: A speech translation dataset with gigaword-scale speech-text pairs.
\newblock \emph{arXiv preprint arXiv:2407.00753}, 2024.

\bibitem[Chorowski et~al.(2019)Chorowski, Weiss, Bengio, and Van Den~Oord]{chorowski2019unsupervised}
Jan Chorowski, Ron~J Weiss, Samy Bengio, and A{\"a}ron Van Den~Oord.
\newblock Unsupervised speech representation learning using wavenet autoencoders.
\newblock \emph{IEEE/ACM transactions on audio, speech, and language processing}, 27\penalty0 (12):\penalty0 2041--2053, 2019.

\bibitem[Défossez et~al.(2022)]{defossez2022encodec}
Alexandre Défossez et~al.
\newblock High fidelity neural audio compression.
\newblock \emph{arXiv preprint arXiv:2210.13438}, 2022.

\bibitem[Fan et~al.(2021)Fan, Bhosale, Schwenk, Ma, El-Kishky, Goyal, Baines, Celebi, Wenzek, Chaudhary, et~al.]{fan2021beyond}
Angela Fan, Shruti Bhosale, Holger Schwenk, Zhiyi Ma, Ahmed El-Kishky, Siddharth Goyal, Mandeep Baines, Onur Celebi, Guillaume Wenzek, Vishrav Chaudhary, et~al.
\newblock Beyond english-centric multilingual machine translation.
\newblock \emph{Journal of Machine Learning Research}, 22\penalty0 (107):\penalty0 1--48, 2021.

\bibitem[Hsu et~al.(2021)]{hsu2021hubert}
Wei-Ning Hsu et~al.
\newblock Hubert: Self-supervised speech representation learning by masked prediction of hidden units.
\newblock \emph{IEEE/ACM Transactions on Audio, Speech, and Language Processing}, 2021.

\bibitem[Ito and Johnson(2017)]{ljspeech17}
Keith Ito and Linda Johnson.
\newblock The lj speech dataset.
\newblock \url{https://keithito.com/LJ-Speech-Dataset/}, 2017.
\newblock Accessed: 2025-05-04.

\bibitem[Jeong et~al.(2021)Jeong, Kim, Cheon, Choi, and Kim]{jeong2021diff}
Myeonghun Jeong, Hyeongju Kim, Sung~Jun Cheon, Byoung~Jin Choi, and Nam~Soo Kim.
\newblock Diff-tts: A denoising diffusion model for text-to-speech.
\newblock \emph{arXiv preprint arXiv:2104.01409}, 2021.

\bibitem[Jia et~al.(2019)]{jia2019translatotron}
Ye~Jia et~al.
\newblock Direct speech-to-speech translation with a sequence-to-sequence model.
\newblock In \emph{Interspeech}, 2019.

\bibitem[Jia et~al.(2022)]{jia2022translatotron2}
Ye~Jia et~al.
\newblock Translatotron 2: Robust direct speech-to-speech translation.
\newblock \emph{arXiv preprint arXiv:2204.02570}, 2022.

\bibitem[Kim et~al.(2020)Kim, Kim, Kong, and Yoon]{kim2020glow}
Jaehyeon Kim, Sungwon Kim, Jungil Kong, and Sungroh Yoon.
\newblock Glow-tts: A generative flow for text-to-speech via monotonic alignment search.
\newblock In \emph{Advances in Neural Information Processing Systems}, 2020.

\bibitem[Kumar et~al.(2015)Kumar, Joshi, Ghosh, and et~al.]{indictts}
Anuj Kumar, Shreyas~S. Joshi, Sreyan Ghosh, and et~al.
\newblock Indictts: A speech corpus for indian languages.
\newblock \url{https://www.iitm.ac.in/donlab/tts/}, 2015.
\newblock Accessed: 2025-05-04.

\bibitem[Lavie et~al.(1997)Lavie, Waibel, Levin, Finke, Gates, Gavalda, Zeppenfeld, and Zhan]{lavie1997janus}
Alon Lavie, Alex Waibel, Lori Levin, Michael Finke, Donna Gates, Marsal Gavalda, Torsten Zeppenfeld, and Puming Zhan.
\newblock Janus-iii: Speech-to-speech translation in multiple languages.
\newblock In \emph{1997 IEEE International Conference on Acoustics, Speech, and Signal Processing}, volume~1, pages 99--102. IEEE, 1997.

\bibitem[Lipman et~al.(2022)]{lipman2022flowmatching}
Yotam Lipman et~al.
\newblock Flow matching for generative modeling.
\newblock \emph{NeurIPS}, 2022.

\bibitem[Liu et~al.(2020)]{liu2020mbart}
Yinhan Liu et~al.
\newblock Multilingual denoising pre-training for neural machine translation.
\newblock \emph{Transactions of the Association for Computational Linguistics}, 2020.

\bibitem[Nakamura et~al.(2006)Nakamura, Markov, Nakaiwa, Kikui, Kawai, Jitsuhiro, Zhang, Yamamoto, Sumita, and Yamamoto]{nakamura2006atr}
Satoshi Nakamura, Konstantin Markov, Hiromi Nakaiwa, Gen-ichiro Kikui, Hisashi Kawai, Takatoshi Jitsuhiro, J-S Zhang, Hirofumi Yamamoto, Eiichiro Sumita, and Seiichi Yamamoto.
\newblock The atr multilingual speech-to-speech translation system.
\newblock \emph{IEEE Transactions on Audio, Speech, and Language Processing}, 14\penalty0 (2):\penalty0 365--376, 2006.

\bibitem[Nichol and Dhariwal(2021)]{nichol2021improved}
Alexander~Quinn Nichol and Prafulla Dhariwal.
\newblock Improved denoising diffusion probabilistic models.
\newblock In \emph{International conference on machine learning}, pages 8162--8171. PMLR, 2021.

\bibitem[Popov et~al.(2021)Popov, Vovk, Gogoryan, Sadekova, and Kudinov]{popov2021grad}
Vadim Popov, Ivan Vovk, Vladimir Gogoryan, Tasnima Sadekova, and Mikhail Kudinov.
\newblock Grad-tts: A diffusion probabilistic model for text-to-speech.
\newblock In \emph{International conference on machine learning}, pages 8599--8608. PMLR, 2021.

\bibitem[Ren et~al.(2019)Ren, Ruan, Tan, Qin, Zhao, Zhao, and Liu]{ren2019fastspeech}
Yi~Ren, Yangjun Ruan, Xu~Tan, Tao Qin, Sheng Zhao, Zhou Zhao, and Tie-Yan Liu.
\newblock Fastspeech: Fast, robust and controllable text to speech.
\newblock \emph{Advances in neural information processing systems}, 32, 2019.

\bibitem[Rombach et~al.(2022)Rombach, Blattmann, Lorenz, Esser, and Ommer]{rombach2022high}
Robin Rombach, Andreas Blattmann, Dominik Lorenz, Patrick Esser, and Bj{\"o}rn Ommer.
\newblock High-resolution image synthesis with latent diffusion models.
\newblock In \emph{Proceedings of the IEEE/CVF conference on computer vision and pattern recognition}, pages 10684--10695, 2022.

\bibitem[Song et~al.(2020{\natexlab{a}})Song, Meng, and Ermon]{song2020denoising}
Jiaming Song, Chenlin Meng, and Stefano Ermon.
\newblock Denoising diffusion implicit models.
\newblock \emph{arXiv preprint arXiv:2010.02502}, 2020{\natexlab{a}}.

\bibitem[Song et~al.(2020{\natexlab{b}})Song, Sohl-Dickstein, Kingma, Kumar, Ermon, and Poole]{song2020score}
Yang Song, Jascha Sohl-Dickstein, Diederik~P Kingma, Abhishek Kumar, Stefano Ermon, and Ben Poole.
\newblock Score-based generative modeling through stochastic differential equations.
\newblock \emph{arXiv preprint arXiv:2011.13456}, 2020{\natexlab{b}}.

\bibitem[Team(2022)]{nllb2022}
NLLB Team.
\newblock No language left behind: Scaling human-centered machine translation.
\newblock \emph{arXiv preprint arXiv:2207.04672}, 2022.

\bibitem[Wang et~al.(2024)]{wang2024vallex}
Chengyi Wang et~al.
\newblock Vall-e x: Zero-shot cross-lingual speech synthesis with discrete codes.
\newblock \emph{arXiv preprint arXiv:2411.14453}, 2024.

\bibitem[Yang and He(2022)]{yang2022cross}
Jingzhou Yang and Lei He.
\newblock Cross-lingual text-to-speech using multi-task learning and speaker classifier joint training.
\newblock \emph{arXiv preprint arXiv:2201.08124}, 2022.

\bibitem[Zhang et~al.(2023{\natexlab{a}})Zhang, Rao, and Agrawala]{zhang2023adding}
Lvmin Zhang, Anyi Rao, and Maneesh Agrawala.
\newblock Adding conditional control to text-to-image diffusion models.
\newblock In \emph{Proceedings of the IEEE/CVF international conference on computer vision}, pages 3836--3847, 2023{\natexlab{a}}.

\bibitem[Zhang et~al.(2023{\natexlab{b}})Zhang, Weiss, Chun, Wu, Chen, Skerry-Ryan, Jia, Rosenberg, and Ramabhadran]{zhang2023multilingual}
Yu~Zhang, Ron~J Weiss, Byungha Chun, Yonghui Wu, Zhifeng Chen, Russell John~Wyatt Skerry-Ryan, Ye~Jia, Andrew~M Rosenberg, and Bhuvana Ramabhadran.
\newblock Multilingual speech synthesis and cross-language voice cloning, February~14 2023{\natexlab{b}}.
\newblock US Patent 11,580,952.

\bibitem[Zhang et~al.(2023{\natexlab{c}})Zhang, Zhou, Wang, Chen, Wu, Liu, Chen, Liu, Wang, Li, et~al.]{zhang2023speak}
Ziqiang Zhang, Long Zhou, Chengyi Wang, Sanyuan Chen, Yu~Wu, Shujie Liu, Zhuo Chen, Yanqing Liu, Huaming Wang, Jinyu Li, et~al.
\newblock Speak foreign languages with your own voice: Cross-lingual neural codec language modeling.
\newblock \emph{arXiv preprint arXiv:2303.03926}, 2023{\natexlab{c}}.

\end{thebibliography}

\end{document}